\documentclass[letter]{article}

\usepackage[margin=1in]{geometry} 

\usepackage{amsmath}
\usepackage{amsthm}
\usepackage{amssymb}

\usepackage[utf8]{inputenc}
\usepackage{hyperref}
\hypersetup{
	unicode,
	pdfauthor={Ahmad Esmaeili, Zahra Ghorrati, Eric T. Matson},
	pdfproducer={LaTeX},
	pdfcreator={pdflatex}
}

\usepackage{tikz}
\usepackage{amsfonts}
\usepackage{forest}
\usepackage{hf-tikz}
\usetikzlibrary{tikzmark}
\usepackage[ruled, linesnumbered, algo2e]{algorithm2e}
\usepackage{algpseudocode}
\usepackage{subcaption}
\usepackage{booktabs}
\usepackage{pgfplots}
\usepgfplotslibrary{fillbetween}
\usepgfplotslibrary{groupplots}
\usepackage{filecontents}
\usepackage{bm}
\usepackage{threeparttable}
\usepackage{booktabs}

\DeclareMathOperator*{\argmin}{arg\,min}
\DeclareMathOperator*{\mean}{mean}

\SetKwProg{Fn}{Function}{:}{end}
\SetFuncSty{textnormal}
\SetKwComment{Comment}{$\triangleright$\ }{}

\SetCommentSty{mycommfont}

\definecolor{mygreen}{rgb}{0.0, 0.5, 0.0}

\theoremstyle{plain}

\theoremstyle{definition}

\usepackage{graphicx, color}
\graphicspath{{fig/}}

\usepackage{algorithm, algpseudocode} 
\usepackage{mathrsfs} 

\title{Agent-based Collaborative Random Search for Hyper-parameter Tuning and Global Function Optimization}
\author{Ahmad Esmaeili\thanks{Corresponding Author} \and Zahra Ghorrati \and Eric T. Matson}

\date{
	$^1$Department of Computer and Information Technology, Purdue University, West Lafayette, IN 47907 \\ \texttt{\{aesmaei, zghorrat, ematson\}@purdue.edu}\\%
}

\begin{document}
	\maketitle
	
	\begin{abstract}
        Hyper-parameter optimization is one of the most tedious yet crucial steps in training machine learning models. There are numerous methods for this vital model-building stage, ranging from domain-specific manual tuning guidelines suggested by the oracles to the utilization of general-purpose black-box optimization techniques. This paper proposes an agent-based collaborative technique for finding near-optimal values for any arbitrary set of hyper-parameters (or decision variables) in a machine learning model (or general function optimization problem). The developed method forms a hierarchical agent-based architecture for the distribution of the searching operations at different dimensions and employs a cooperative searching procedure based on an adaptive width-based random sampling technique to locate the optima. The behavior of the presented model, specifically against the changes in its design parameters, is investigated in both machine learning and global function optimization applications, and its performance is compared with that of two randomized tuning strategies that are commonly used in practice. According to the empirical results, the proposed model outperformed the compared methods in the experimented classification, regression, and multi-dimensional function optimization tasks, notably in a higher number of dimensions and in the presence of limited on-device computational resources.\\
		
	\end{abstract}

    \section{Introduction}\label{sec:introduction}
Almost all Machine Learning (ML) algorithms comprise a set of hyper-parameters that control their learning process and the quality of their resulting models. The number of hidden units, the learning rate, the mini-batch sizes, etc. in neural networks; the kernel parameters and regularization penalty amount in support vector machines; and maximum depth, samples split criteria, and the number of used features in decision trees are few common hyper-parameter examples that need to be configured for the corresponding learning algorithms. Assuming a specific ML algorithm and a dataset, one can build countless number of models,  each with potentially different performance and/or learning speeds, by assigning different values to the algorithm's hyper-parameters. While they provide ultimate flexibility in using ML algorithms in different scenarios, they also account for most failures and tedious development procedures. Unsurprisingly, there are numerous studies and practices in the machine learning community devoted to the optimization of hyper-parameter. The most straightforward yet difficult approach utilizes expert knowledge to identify potentially better candidates in hyper-parameter search spaces to evaluate and use. The availability of expert knowledge and generating reproducible results are among the primary limitation of such manual searching technique \cite{bergstra2012random}, particularly due to the fact that using any learning algorithm on different datasets likely requires different sets of hyper-parameter values~\cite{kohavi1995automatic}.

Formally speaking, let $\Lambda=\{\bm{\lambda}\}$ denote the set of all possible hyper-parameter value vectors and $\mathcal{X}=\{\mathcal{X}^{(train)}, \mathcal{X}^{(valid)}\}$ be the dataset split into training and validation sets. The learning algorithm with hyper-parameter values vector $\bm{\lambda}$ is a function that maps training dataset $\mathcal{X}^{(train)}$ to model $\mathcal{M}$, i.e. $\mathcal{M}=\mathcal{A}_{\bm{\lambda}}(\mathcal{X}^{(train)})$, and the hyper-parameter optimization problem can be formally written as \cite{bergstra2012random}:
\begin{equation}\label{eq:HOP}
	\bm{\lambda}^{(*)}=\argmin_{\bm{\lambda}\in\Lambda}\mathbb{E}_{x\sim\mathcal{G}_x}\left[\mathcal{L}\left(x; \mathcal{A}_{\bm{\lambda}}(\mathcal{X}^{(train)})\right)  \right] 
\end{equation}
where $\mathcal{G}_x$ and $\mathcal{L}(x;\mathcal{M})$ are respectively the grand truth distribution and the expected loss of applying learning model $\mathcal{M}$ over i.i.d. samples $x$; and $\mathbb{E}_{x\sim\mathcal{G}_x}\left[\mathcal{L}\left(x; \mathcal{A}_{\bm{\lambda}}(\mathcal{X}^{(train)})\right)  \right]$ gives the generalization error for algorithm $\mathcal{A}_{\bm{\lambda}}$. To cope with the inaccessibility of the grand truth in real-world problems, the generalization error is commonly estimated using the cross-validation technique \cite{10.1162/EVCO_a_00069}, leading to the following approximation of the above-mentioned optimization problem:
\begin{equation}\label{eq:HOP_CV}
	\bm{\lambda}^{(*)}\approx\argmin_{\bm{\lambda}\in\Lambda}\mean_{x\in\mathcal{X}^{(valid)}}\mathcal{L}\left(x; \mathcal{A}_{\bm{\lambda}}(\mathcal{X}^{(train)})\right)\equiv\argmin_{\bm{\lambda}\in\Lambda} \Psi(\bm{\lambda})
\end{equation}
where $\Psi(\bm{\lambda})$ is called hyper-parameter response function \cite{bergstra2012random}.

Putting the manual tuning approaches aside, there is a wide range of techniques that use global optimization methods to address the ML hyper-parameter tuning problem. Grid search \cite{montgomery2017design,John94cross-validatedc4.5:}, random search \cite{bergstra2012random}, Bayesian optimization \cite{movckus1975bayesian,mockus2012bayesian,feurer2019hyperparameter}, and evolutionary and population-based optimization \cite{simon2013evolutionary,esmaeili2009adjusting,9504761} are some common class of tuning methodologies that are studied and used extensively by the community. 
In grid search, for instance, every combination of a predetermined set of values in each hyper-parameter is evaluated and the hyper-parameter value vector that minimizes the loss function is selected. For $k$ number of configurable hyper-parameters, if we denote the set of candidate values for the $j$-th hyper-parameter $\lambda_j^{(i)}\in \bm{\lambda}^{(i)}$ by $\mathcal{V}_j$, grid search would evaluate  $T=\Pi_{j=1}^k|\mathcal{V}_j|$ number of trials that can grow exponentially with the increase in the number of configurable hyper-parameters and the quantity of the candidate values for each dimension. This issue is referred to as the curse of dimensionality \cite{bellman1961adaptive} and is the primary reason for making grid search an uninteresting methodology in large-scale real-world scenarios. 
In the standard random search, on the other hand, a set of $b$ uniformly distributed random points in the hyper-parameter search space, $\{\lambda^{(1)}, \dots, \lambda^{(b)}\}\in \Lambda$, are evaluated to select the best candidate. As the number of evaluations only depends on the budget value, $b$, random search does not suffer from the curse of dimensionality, is shown to be more effective than grid search \cite{bergstra2012random}, and is often used as a baseline method. Bayesian optimization, as a global black box expensive function optimization technique, iteratively fits a surrogate model to the available observations, $(\lambda^{(i)},\Psi_{\lambda^{(i)}})$ and then, uses an acquisition function to determine the next hyper-parameter values to evaluate and use in the next iteration \cite{feurer2019hyperparameter,shahriari2015taking}. Unlike grid and random search methods, in which the searching operations can be easily parallelized, the Bayesian method is originally sequential though various distributed versions have been proposed in the literature \cite{https://doi.org/10.48550/arxiv.1902.09992,young2020distributed}. Nevertheless, thanks to its sample efficiency and robustness to noisy evaluations, Bayesian optimization is a popular method in the hyper-parameter tuning of deep-learning models, particularly when the number of configurable hyper-parameters is less than 20 \cite{BOTutorial}. Evolution and population-based global optimization methods, such as genetic algorithms and swarm-based optimization techniques, form the other class of common tuning approaches, in which the hyper-parameters configurations are improved over multiple generations generated by local and global perturbations \cite{friedrichs2005evolutionary,esmaeili2009adjusting,9504761}. Population-based methods are embarrassingly parallel \cite{loshchilov2016cma} and, similar to grid and random search approaches, the evaluations can be distributed over multiple machines. 

Multi-Agent Systems (MAS) and agent-based technologies, when applied to machine learning and data mining, bring about scalability and autonomy, and facilitate the decentralization of learning resources and the utilization of strategic and collaborative learning models \cite{ryzko2020modern,esmaeili2022hamlet,esmaeili2022paams}. Agnet-based machine learning and collaborative hyper-parameter tuning are not new to this paper and have previously been studied in the literature. The research reported in \cite{bardenet2013collaborative} is among the noteworthy contributions, which proposes a surrogate-based collaborative tuning technique incorporating the experience achieved from previous experiments. To put it simply, this model performs simultaneous configuration of the same hyper-parameters over multiple datasets and employs the gained information in all subsequent tuning problems. Auto-Tuned Models (ATM) \cite{swearingen2017atm} is a distributed and collaborative system that automates hyper-parameter tuning and classification model selection procedures. At its core, ATM utilizes Conditional Parameter Tree (CPT), in which a learning method is placed at the root and its children are the method's hyper-parameters, to represent the hyper-parameter search space. Different tunable subsets of hyper-parameter nodes in CPT are selected during model selection, and assigned to a cluster of workers to be configured. Koch et al. \cite{koch2018autotune} introduced Autotune as a derivative-free hyper-parameter optimization framework. Being composed of a hybrid and extendable set of solvers, this framework concurrently runs various searching methods potentially distributed over a set of workers to evaluate objective functions and provide feedback to the solvers. Autotune employs an iterative process during which all the points that have already been evaluated are exchanged  with the solvers to generate new sets of points to evaluate. In learning-based settings, the work reported in \cite{iranfar2021multi} used Mutli-Agent Reinforcement Learning (MARL) to optimize the hyper-parameters of deep convolutional neural networks (CNN). The suggested model splits the design space into sub-spaces and devotes each agent to tuning the hyper-parameters of a single network layer using Q-learning. Parker-Holder et al. \cite{parker2020provably} presented the Population-Based Bandit (PB2) algorithm, which efficiently directs the searching operation of hyper-parameters in Reinforcement Learning using a probabilistic model. In PB2, a population of agents is trained in parallel, and their performance is monitored on a regular basis. An underperforming agent's network weights are replaced with those of a better-performing agent, and its hyper-parameters are tuned using Bayesian optimization. 

In continuation to our recent generic collaborative optimization model \cite{esmaeili2022paams}, this paper delves into the design of a multi-level agent-based distributed random search technique that can be used for both hyper-parameter tuning and general-purpose black-box function optimization. The proposed methods, at its core, forms a tree-like structure comprising a set of interacting agents that, depending on their position in the hierarchy, focus on tuning/optimizing a single hyper-parameter/decision variable using a biased hyper-cube random sampling technique,  or aggregating the results and facilitating collaborations based on the gained experience of other agents. The rationales behind choosing random search as the core tuning strategy of the agents include, but are not limited to, its intrinsic distributability, acceptable performance in practice, and not requiring differentiable objective functions. Although the parent model in \cite{esmaeili2022paams} does not impose any restrictions on the state and capabilities of the agents, this paper assumes homogeneity in the sense that the tuner/optimizer agents use the same mechanism for their assigned job. With that said, the proposed method is analyzed in terms of its design parameters, and the empirical results from conducted ML classification and regression tasks, as well as, various multi-dimensional functions optimization problems demonstrate that the suggested approach not only outperforms the underlying random search methodologies under the same deployment conditions but also provide a better-distributed solution in the presence of limited computational resources. 

The remainder of this paper is organized as follows: section~\ref{sec:proposed} dissects the proposed agent-based random search method; section~\ref{sec:results} presents the details of used experimental ML and function optimization settings, and discusses the performance of the proposed model under different scenarios; and finally, section~\ref{sec:conclusion} concludes the paper and provides future work suggestions. 

    \section{Methodology}\label{sec:proposed}
This section dissects the proposed agent-based hyperparameter tuning and black-box function optimization approach. To help with a clear understanding of the proposed algorithms, this section begins by providing the preliminaries and introducing the key concepts and then presents the details of the agent-based randomized search algorithms accompanied by hands-on examples whenever needed.

\subsection{Preliminaries}\label{sec:prelim}
An agent, as the first-class entity in the proposed approach, might play different roles depending on its position in the system. As stated before, this paper is using hierarchical structures to coordinate the agents in the system and hence, it defines two agent types: (1) \textit{internals}, which play the role of result aggregators and collaboration facilitators in connection with their subordinates; and (2) \textit{terminals}, which, implementing a single-variable randomized searching algorithm, are the actual searchers/optimizers positioned at the bottom-most level of the hierarchy. Assuming $G$ to be the set of all agents in the system and the root of the hierarchy to be at level 0, this paper uses $g_{\lambda_i}^{l}$ ( $G_{\bm{\lambda_j}}^{l}$) to denote the agent (the set of agents) at level $l$ of the hierarchy that are specialized in tuning hyper-parameter $\lambda_i$ (hyper-parameter set $\bm{\lambda_j}$) respectively, where $\bm{\lambda_j}\subseteq \bm{\lambda}$ and $g_{\lambda_i}^{l+1} \in G_{\bm{\lambda_j}}^{l}$ iff. $\lambda_i \in \bm{\lambda_j}$.

As denoted above, the hyper-parameters that the agents represent determine their position in the hierarchy. let $\mathcal{A}_{\bm{\lambda}=\{\lambda_2,\lambda_2, \ldots \lambda_n\}}$ be the ML algorithms for which we intend to tune the hyper-parameters. As the tuning process might not target the entire hyper-parameter set of the algorithm, the proposed method divides the set into two \textit{objective} and \textit{fixed} disjoint subsets, which respectively denoted by $\bm{\lambda_o}$ and $\bm{\lambda_f}$ refer to the hyper-parameter sets that we intend to tune and the ones we need to keep fixed. Formally, that is $\bm{\lambda}=\bm{\lambda_o}\cup\bm{\lambda_f}$ and $\bm{\lambda_o}\cap\bm{\lambda_f}=\emptyset$. The paper further assumes two types of objective hyper-parameters: (1) \textit{primary} hyper-parameters, denoted by $\bm{\hat{\lambda}_o}$, which comprise the main targets of the corresponding tuners (agents); and (2) \textit{subsidiary} hyper-parameters, denoted by $\bm{\hat{\lambda}'_o}$, which include the ones whose values are set by the other agents to help limit the searching space. These two sets are complements of each other, i.e., $\bm{\hat{\lambda}'_o}=\bm{\lambda_o}-\bm{\hat{\lambda}_o}$, and the skill of an agent is determined by the primary objective set, $\bm{\hat{\lambda}_o}$, it represents. With that said, for all terminal agents in the hierarchy, we have $|\bm{\hat{\lambda}_o}|=1$, where $|\dots|$ denotes the set cardinality.

The agents of a realistic MAS are susceptible to various limitations that are imposed by their environment and/or computational resources. This paper, due to its focus on the decentralization of the searching process, foresees two limitation aspects for the agents: (1) the maximum number of concurrent connections, denoted by $c$, that an agent can manage; (2) the number of concurrent processes, called \textit{budget} and denoted by $b$, that the agent can execute and handle. In the proposed method, $c>1$ determines the maximum number of subordinates (children) that an internal agent can have. The budget, $b\geq 1$, on the other hand, put a restriction on the maximum number of parallel evaluations that an agent can perform in the step of searching for the optima.

Communications play a critical role in all MAS, including the agent-based method proposed in this paper. For all intra-system, i.e. between any two agents, and inter-system, i.e. between an agent and a user, interactions, the suggested method uses the following tuple-based structure for the queries: 
\begin{equation}\label{query}
	\left<\mathcal{A}_{\bm{\lambda}},\{\bm{\hat{\lambda}_o},\bm{\hat{\lambda}'_o}, \bm{\lambda_f}\}, \bm{\mathcal{V}}, \{\mathcal{X}^{(train)},\mathcal{X}^{(valid)}\}, \mathcal{L}\right>
\end{equation}
where, $\bm{\mathcal{V}}=\{(\lambda_i,v_i)\}_{1\leq i\leq n}$ denotes the set containing the candidate values for all hyper-parameter, and the remaining notations are as defined in equation~\ref{eq:HOP}. Based on what discussed before, it is clear that $|\bm{\lambda_f}|\leq|\bm{\mathcal{V}}|\leq\bm{\lambda}$. 

\subsection{Agent-based Randomized Searching Algorithm}

The high-level abstract view of the proposed approach is composed of two major steps: (1) distributedly building the hierarchical MAS; and (2) performing the collaborative searching process through vertical communications in the hierarchy. The sections that follow go into greater detail about these two stages.

\subsubsection{Distributed Hierarchy Formation}
\begin{figure}
	\centering
    \includegraphics[width=\textwidth]{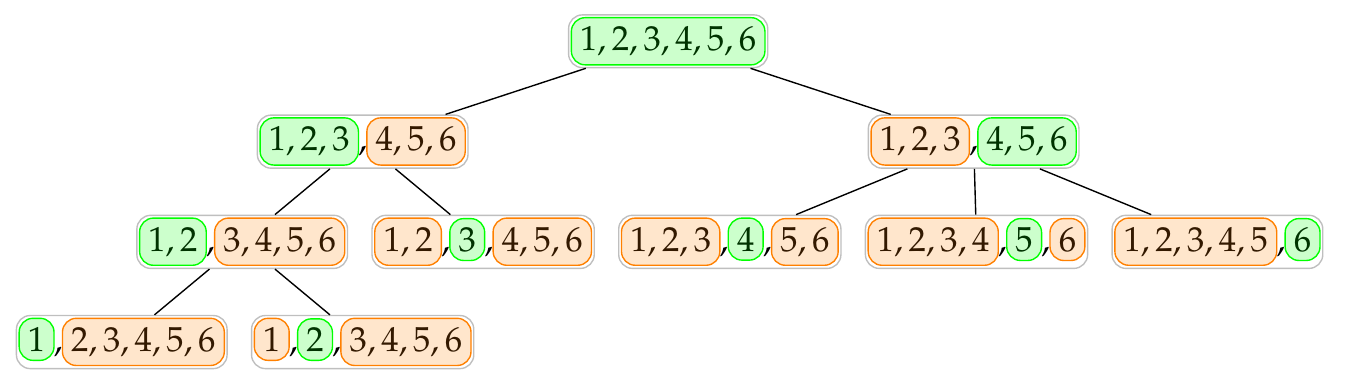}
	\caption{Hierarchical structure built for $\bm{\lambda_o}=\{\lambda_1, \lambda_2, \lambda_3, \lambda_4, \lambda_5, \lambda_6\}$, where the primary and complementary hyper-parameters of each node are respectively highlighted in green and orange, and the labels are the indexes of $\lambda_i$.}
	\label{fig:hyper-hier}
\end{figure}

As for the first step, each agent, $t$, divides the primary objective hyper-parameter set of the query it receives, i.e. $\bm{\hat{\lambda}_o}$, into $c_t>1$ number of subsets, for each of which, the system initiates a new agent to handle. This process continues recursively until there is only one hyper-parameter in the primary objective set, i.e. $|\bm{\hat{\lambda}_o}|=1$, which is assigned to a terminal agent.
Figure~\ref{fig:hyper-hier} provides an example hierarchy resulted from the recursive division of primary objective set $\bm{\lambda_o}=\{\lambda_1, \lambda_2, \lambda_3, \lambda_4, \lambda_5, \lambda_6\}$. For the sake of clarity, we have used the indexes of the hyper-parameters as the labels of the nodes in the hierarchy, and the green and orange colors are employed to highlight the members of the $\color{mygreen}{\bm{\hat{\lambda}_o}}$ and $\color{orange}{\bm{\hat{\lambda}'_o}}$ sets respectively. Regarding the maximum number of concurrent connections that the agents can handle in this example, it is assumed for all agents $c=2$, except for the rightmost agent in the second level of the hierarchy, for which $c=3$. It is worth emphasizing that at the beginning of the process, when the tuning query is received from the user, we have $\bm{\hat{\lambda}_o}=\bm{\lambda_o}$ and $\bm{\hat{\lambda}'_o=\emptyset}$, which is the reason for the all green node of the root node in this example. 

Algorithm~\ref{alg:form-hier} presents the details of the process. We have chosen self-explanatory names for the functions and variables and provided comments wherever they are required to improve clarity. In this algorithm, the function \textsc{PrepareResources} in line~\ref{ln:resource} prepares the data and computational resources for the newly built/assigned terminal agent. Such resources are used for training, validation, and tuning process. The function \textsc{SpawnOrConnect} in line~\ref{ln:spawn} creates a subordinate agent that represents ML algorithm $\mathcal{A}_\lambda$ and expected loss function $\mathcal{L}$. This is done by either creating a new agent or connecting to an existing idle one if the resources are reused. Two functions \textsc{PrepareFeedback} and \textsc{Tune} in lines~\ref{ln:feed} and \ref{ln:tune}, respectively, are called when the structure formation process is over and the root agent initiates the tuning process in the hierarchy. Later, these two functions are discussed in more detail.

\begin{algorithm}
	\DontPrintSemicolon
	\SetKwFunction{FStart}{\textsc{Start}}
	\SetKwFunction{FTune}{\textsc{Tune}}
	\caption{Distributed formation of the hierarchical agent-based hyper-parameter tuning structure.}
	\label{alg:form-hier}
	\Fn{\FStart{$\left<\mathcal{A}_{\bm{\lambda}},\{\bm{\hat{\lambda}_o},\bm{\hat{\lambda}'_o}, \bm{\lambda_f}\},\bm{\mathcal{V}}, \{\mathcal{X}^{(train)},\mathcal{X}^{(valid)}\}, \mathcal{L}\right>$}}{
		\eIf(\Comment*[f]{agent is terminal}){$|\bm{\hat{\lambda}_o}|=1$}{
			$\mathcal{R}\gets$\Call{PrepareResources}{$\left<\{\bm{\hat{\lambda}_o},\bm{\hat{\lambda}'_o}, \bm{\lambda_f}\}, \{\mathcal{X}^{(train)},\mathcal{X}^{(valid)}\}\right>$}\;\label{ln:resource}
			\Call{Inform}{$Parent,\mathcal{R}$}\Comment*[f]{informs the parent agent}
		}
		(\Comment*[f]{agent is internal ($|\bm{\hat{\lambda}_o}|>1$)}){
			$k\gets \min(c_{my}, |\bm{\hat{\lambda}_o}|)$\Comment*[f]{the number of children}\\
			\For{$i\gets1$ \KwTo $k$}{
				$G_i\gets$ \Call{SpawnOrConnect}{$\mathcal{A}_{\bm{\lambda}}, \mathcal{L}$}\;\label{ln:spawn}
				$\bm{\hat{\lambda}_{o_i}}\gets$\Call{Divide}{$\bm{\hat{\lambda}_o}$,$i$, $k$}\Comment*[f]{the $i^{\text{th}}$ unique devision}\\
				$\bm{\hat{\lambda}'_{o_i}}\gets(\bm{\hat{\lambda}_{o}} - \bm{\hat{\lambda}_{o_i}})\cup\bm{\hat{\lambda}'_{o}}$\;
				$\mathcal{R}_i\gets$\Call{Ask}{$G_i$, \textsc{Start}, $\left<\mathcal{A}_{\bm{\lambda}},\{\bm{\hat{\lambda}_{o_i}},\bm{\hat{\lambda}'_{o_i}}, \bm{\lambda_f}\}, \bm{\mathcal{V}},\{\mathcal{X}^{(train)},\mathcal{X}^{(valid)}\}, \mathcal{L}\right>$}	
			}
			$\mathcal{R}\gets$\Call{Aggregate}{$\{\mathcal{R}_i\}_i$}\Comment*[f]{combines children's answers}\\
			\eIf{Parent $\ne\emptyset$}{
				\Call{Inform}{$Parent, \mathcal{R}$}\;
			}{
				$\mathcal{F}\gets\Call{PrepareFeedback}{\mathcal{R}, \bm{\mathcal{V}}}$\;\label{ln:feed}
				\Call{Tune}{$\mathcal{F}$}\Comment*[f]{initiates the tuning process}\label{ln:tune}\\
			}
		}
	
	}
\end{algorithm}

\subsubsection{Collaborative tuning Process}
The collaborative tuning process is conducted through a series of vertical communications in the built hierarchy. Initiated by the root agent, as explained in the previous section, the \textsc{Tune} request is propagated to all of the agents in the hierarchy. As for the internal agents, the request will be simply passed down to the subordinates as they arrive. As for the terminal agents, on the other hand, the request launches the searching process in the sub-space specified by the parent. The flow of the results will be in an upward direction with a slightly different mechanism. As soon as a local optimum is found by a terminal agent, it will be sent up to the parent agent. Having waited for the results to be collected from all of its subordinates, the parent aggregates them together and passes the combined result to its own parent. This process continues until it reaches the root agent where the new search guidelines are composed for the next search round. 

Algorithm~\ref{alg:tune} presents the details of the iterated collaborative tuning process, which might be called by both terminal and internal agents. When it is called by a terminal agent, it initiates the searching operation for the optima of the hyper-parameter that the agent represents and informs the result to its parent. Let $g_{\lambda_j}^l$ be the terminal agent concentrating on tuning hyper-parameter $\lambda_j$. As it can be seen in line~\ref{ln:run}, the result of the search will be a single-item set composed of the identifier of the hyper-parameter, i.e. $\lambda_j$, the set $\bm{\mathcal{V}_j}^{(*)}$ containing the coordinate of the best candidate agent $g_{\lambda_j}^l$ has found, and the response function value for that best candidate, i.e. $\Psi^{(*)}_j$. An internal agent running this procedure, on the other hand, merely passes the tuning request to the subordinates and waits for their search results (line~\ref{ln:askchildren} of the algorithm). Please note that this asking operation comprises filtering operation on set $\bm{\mathcal{F}}$. That is, the subordinate will receive a subset $\bm{\mathcal{F}}_i\subset\bm{\mathcal{F}}$ that only includes starting coordinates for the terminal agents reachable through that agent. Having collected all of the results from its subordinates, the internal agent aggregates them by simply joining the result sets and informs its own parent, in case it is not the root agent. This process is executed recursively until the aggregated results reach the root agent of the hierarchy. Depending on whether the stopping criteria of the algorithm are reached, the root prepares feedback to initiate the next tuning iteration or a report detailing the results. The collaboration between the agents is conducted implicitly through the feedback that the root agent provides for each terminal agent based on the results it has gathered in the previous iteration. As presented in line~\ref{ln:feed} of the algorithm, this feedback basically determines the coordinates of the position where the terminal agents should start their searching operation. It should be noted that the \texttt{argmin} function in this operation is due to employing loss function $\mathcal{L}$ as a metric to evaluate the performance of an ML model. For performance measures in which maximization is preferred, such as in \textit{accuracy}, this operation needs to be replaced by \texttt{argmax} accordingly. 

\begin{algorithm}
	\DontPrintSemicolon
	\SetKwFunction{FStart}{\textsc{Start}}
	\SetKwFunction{FTune}{\textsc{Tune}}
	\caption{Iterated collaborative tuning procedure.}
	\label{alg:tune}
	\Fn{\FTune{$\bm{\mathcal{F}}$}}{
		\eIf(\Comment*[f]{terminal agent agent}){$Children=\emptyset$}{
			$\{(\lambda_j,\bm{\mathcal{V}}^{(*)}_j, \Psi^{(*)}_j)\}\gets$\Call{RunTuningAlgorithm}{$\bm{\mathcal{F}}=\bm{\mathcal{V}}$}\label{ln:run}\\
			\Call{Inform}{$Parent,\{(\lambda_j,\bm{\mathcal{V}}^{(*)}_j, \Psi^{(*)}_j)\}$}
		}{
			\ForEach{$G_i^{l+1}\in{\normalfont Children}$}{
				$\bm{\mathcal{R}}^{(*)}_i\gets$\Call{Ask}{$G_i^{l+1}$, \textsc{Tune}, $\bm{\mathcal{F}_i}\subset\bm{\mathcal{F}}$}\label{ln:askchildren}\Comment*[f]{$\bm{\mathcal{R}}^{(*)}_i=\{(\lambda_k,\bm{\mathcal{V}}^{(*)}_k, \Psi^{(*)}_k)\}_k$}\;
			}
			$\bm{\mathcal{R}}^{(*)}\gets\bigcup\limits_{G_i^{l+1}\in Children}\bm{\mathcal{R}}^{(*)}_i $
   \Comment*[f]{aggregates results}\\
			\eIf(\Comment*[f]{non-root internal agent}){$Parent\ne\emptyset$}{
				\Call{Inform}{$Parent,\bm{\mathcal{R}}^{(*)}$}\;
			}{
				\eIf{\textsc{ShouldStop}{\normalfont ($StopCriteria$)$\ne \text{True}$}}{\label{ln:stop}
                    $\bm{\mathcal{F}}\gets
     \left\{(\lambda_i, \bm{\mathcal{V}}_j); j = \argmin\limits_{1\leq k\leq n} \Psi^{(*)}_k\right\}_{1\leq i\leq n}$\Comment*[f]{prepares feedback}\;
					\Call{Tune}{$\bm{\mathcal{F}}$}\Comment*[f]{initiates next tuning iteration}\\
				}{
					\Call{PrepareReport}{$\bm{\mathcal{R}}^{(*)}$}\Comment*[f]{reports final result}
				}
			}
			
		}
	}
\end{algorithm}

The details of the tuning function that each terminal agent runs in line~\ref{ln:run} of algorithm~\ref{alg:tune} to tune a single hyper-parameter are presented in algorithm~\ref{alg:strategy}. As its input, this function receives a coordinate that the agent $g_{\lambda_i}^l$ will use as its starting point in the searching process. The received argument together with $b$ additional coordinates that the agent generates randomly are stored in the set of candidate $\bm{\mathcal{C}}$. Accordingly, $\bm{\mathcal{C}}[c]$ and $\bm{\mathcal{C}}[c](\lambda_i)$ refer to the $c$-th coordinate in the set and the value assigned to hyper-parameter $\lambda_i$ of that coordinate respectively. Moreover, please recall from section~\ref{sec:prelim} that $b$ denotes the evaluation budget of a terminal agent. The terminal agents in the proposed method employ a slot-based uniform random sampling to explore the search space. Formally, let $\bm{\mathcal{E}}=\{\epsilon_{\lambda_1}, \epsilon_{\lambda_2},\dots, \epsilon_{\lambda_n}\}$ be a set of real values that each agent utilizes for each hyper-parameter to control the size of slots in any iteration. Similarly, let $\bm{s}=\{s_{\lambda_1},s_{\lambda_2},\dots,s_{\lambda_n}\}$ specify the coordinate of the position that an agent starts its searching operation in any iteration. To sample $b$ random values in the domain $\mathbb{D}_{\lambda_j}$ of any arbitrary hyper-parameter $\lambda_j$, the agent will generate one uniform random value in range
\begin{equation}
    \mathcal{R}=\left[\max(\inf\mathbb{D}_{\lambda_j}, s_{\lambda_j}-\epsilon_{\lambda_j}), \min(\sup\mathbb{D}_{\lambda_j}, s_{\lambda_j}+\epsilon_{\lambda_j}))\right]
\end{equation} and $b-1$ random values in $\mathbb{D}_{\lambda_j} - \mathcal{R}$ by splitting it into $b-1$ slots and choosing one uniform random value in each slot (lines~\ref{ln:onerand} and \ref{ln:remrands} of the algorithm). The generation of the uniform random values is done by calling function \textsc{UniformRand}($A_1, A_2, A_3$). This function divides range $A_1$ into $A_2$ equal-sized slots and returns the uniform random value generated in the $A_3$-th slot. As it can be seen in line~\ref{ln:otherdoms} of the algorithm, the agent employs the same function to generate one and only one value per each element in its subsidiary objective hyper-parameter set $\bm{\hat{\lambda}'_o}$.  

The slot width parameter set $\bm{\mathcal{E}}$ is used to control the exploration behavior of the agent around the starting coordinates in the search space. For instance, for any arbitrary hyper-parameter $\lambda_i$, very small values of $\epsilon_{\lambda_i}$ emphasize generating candidates in the close vicinity of the starting position. On the other hand, larger values of $\epsilon_{\lambda_i}$ decrease the chance that the generated candidate be close to the starting position. In the proposed method, the agents adjust $\bm{\mathcal{E}}$ adaptively. To put it formally, each agent starts the tuning process with the pre-specified value set $\bm{\mathcal{E}}^{(0)}$, and assuming that $\bm{\mathcal{C}}^{(*)}$ denotes the best candidate that the agent has found in the previous iteration, the width parameter set $\bm{\mathcal{E}}$ in iteration $i$ is updated as follows: 
\begin{equation}
   \bm{\mathcal{E}}^{(i)} = 
        \begin{cases}
            \bm{\Delta}\odot\bm{\mathcal{E}}^{(i-1)} & \text{If } \bm{\mathcal{V}}=\bm{\mathcal{C}}^{(*)}\\
            \bm{\mathcal{E}}^{(i-1)} & \text{otherwise}
        \end{cases}
\end{equation}
where $\Delta=\{\delta_{\lambda_1}, \delta_{\lambda_2},\dots, \delta_{\lambda_n}\}$ denotes the scaling changes to apply to the width parameters, and $\odot$ denotes the element-wise multiplication operator. As the paper discusses in section~\ref{sec:results}, despite the generic definitions provided here for futuristic extensions, using the same scaling value for all primary hyper-parameters has led to satisfactory results in our experiments. 

\begin{algorithm}
	\DontPrintSemicolon
	\SetKwFunction{FRTune}{\textsc{RunTuningAlgorithm}}
	\caption{A terminal agent's randomized tuning process.}
	\label{alg:strategy}
	\Fn{\FRTune{$\bm{\mathcal{F}}=\bm{\mathcal{V}}=\{(\lambda_m,v_m)\}_{1\leq m\leq n}$}}{
		$\bm{\mathcal{C}}[0]\gets \bm{\mathcal{V}}$\;
        $\mathcal{R}_{\lambda_i}\gets\left[\max(\inf\mathbb{D}_{\lambda_i}, v_i-\epsilon_{\lambda_i}), \min(\sup\mathbb{D}_{\lambda_i}, v_i+\epsilon_{\lambda_i}))\right]$\;
        
		\For{$c\gets 1$ \KwTo $c=b$}{
            \eIf(\Comment*[f]{the first sample for $\bm{\hat{\lambda}_o}=\{\lambda_i\}$}){$c=1$}{
				$\bm{\mathcal{C}}[c](\lambda_i)\gets$\Call{UniformRand}{$\mathcal{R}_{\lambda_i}, 1, 1$}\label{ln:onerand}\;
			}(\Comment*[f]{the remaining samples for $\bm{\hat{\lambda}_o}=\{\lambda_i\}$}){
                $\bm{\mathcal{C}}[c](\lambda_i)\gets$\Call{UniformRand}{$\mathbb{D}_{\lambda_i}-\mathcal{R}_{\lambda_i}, b-1, c-1$}\label{ln:remrands}\;
            }
                \ForAll{$\lambda_k\in\bm{\hat{\lambda}'_o}$}{
                    $\mathcal{R}_{\lambda_k}\gets\left[\max(\inf\mathbb{D}_{\lambda_k}, v_k-\epsilon_{\lambda_k}), \min(\sup\mathbb{D}_{\lambda_k}, v_k+\epsilon_{\lambda_k}))\right]$\;

                    $\bm{\mathcal{C}}[c](\lambda_k)\gets$\Call{UniformRand}{$\mathcal{R}_{\lambda_k}, 1, 1$}\label{ln:otherdoms}\;
                    
			}
		}
		$\bm{\mathcal{C}}^{(*)}\gets\argmin\limits_{0\leq j\leq b} \Psi(\bm{\mathcal{C}}[j])$\;
		\Return $\{(\lambda_i, \bm{\mathcal{C}}^{(*)}, \Psi(\bm{\mathcal{C}}^{(*)})\}$\;
	}
\end{algorithm}

\begin{figure}
	\centering
	\includegraphics[width=.9\textwidth]{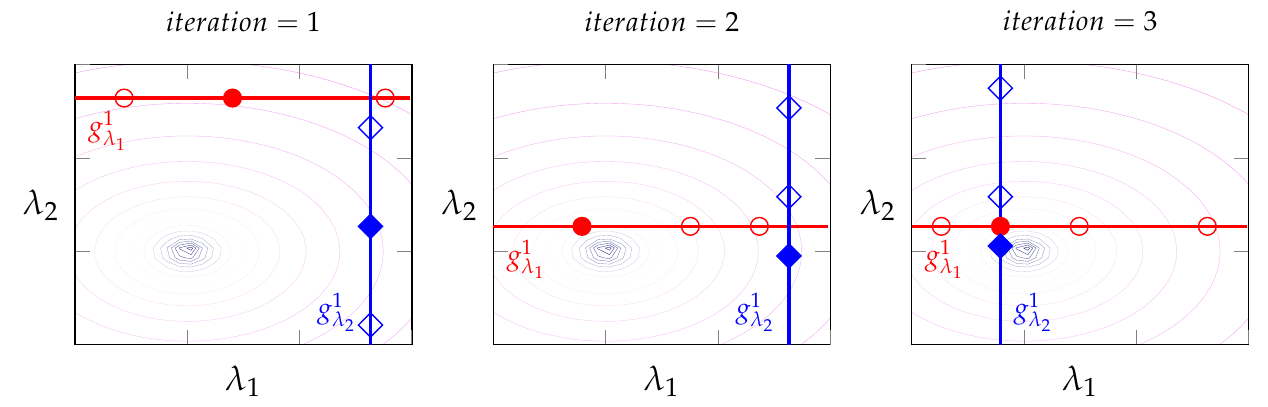}
	\caption{A toy example demonstrating 3 iterations of running the proposed method for tuning two hyper-parameters $\lambda_1$, and $\lambda_2$ using terminal agents $g_{\lambda_1}^1$ and $g_{\lambda_2}^1$ respectively. It is assumed that for each agent $b=3$.}
	\label{fig:tune-ex}
\end{figure}

To better understand the suggested collaborative randomized tuning process of agents, an illustrative example is depicted in figure~\ref{fig:tune-ex}. In this figure, each agent is represented by a different color, and the best candidate that each agent finds at the end of each iteration is shown by a filled shape. Moreover, we have assumed that the value of the loss function becomes smaller as we move inwards in the depicted contour lines, and to prevent any exploration in the domain of the subsidiary hyper-parameters, we have set $\mathcal{E}=\{\epsilon_{\lambda_1}=\frac{1}{6}, \epsilon_{\lambda_2}=0\}$ and $\mathcal{E}=\{\epsilon_{\lambda_1}=0, \epsilon_{\lambda_2}=\frac{1}{6}\}$ for agents $g^1_{\lambda_1}$ and $g^1_{\lambda_2}$ respectively, assuming that the domain size of each hyper-parameter is 1 and $b=3$. In Iteration 1,  both agents start at the top right corner of the search space and are able to find candidates that yield lower loss function values than the starting coordinate. For iteration 2, the starting coordinate of each agent is set to the coordinate of the best candidate found by all agents in the previous iteration. As the best candidate was found by agent $g^1_{\lambda_2}$, we only see the change in the searching direction of the red agent, i.e. $g^1_{\lambda_2}$. The winner agent at the end of this iteration is agent $g^1_{\lambda_1}$, hence we do not see any change to its searching direction in iteration 3. Please note that having 4 circles for agent $g^1_{\lambda_1}$ in the last depicted iteration is because of showing the starting coordinate, which happens to stay the best candidate, in this iteration. It is also worth emphasizing that the starting coordinates are not evaluated again by the agents, as they have already been accompanied by their corresponding response values from the previous iterations.

    \section{Results and Discussion}\label{sec:results}

This section dissects the performance of the proposed method in more detail. It begins with the computational complexity of the technique, and then, provides empirical results on both machine learning and general function optimization tasks.

\subsection{Computational Complexity}\label{sec:complx}

Forming the hierarchical structure and conducting the collaborative searching process are the two major stages of the proposed method. These stages need to be conducted in sequence, and the rest of this section investigates the complexity of each step separately and in relation to one another. 

Regarding the structural formation phase of the suggested method, the shape of the hierarchy depends on the maximum number of connections that each agent can handle; the fewer the number of manageable concurrent connections, the deeper the resulting hierarchy. Using the same notations presented in section~\ref{sec:prelim} and assuming the same $c>1$ for all agents, the depth of the formed hierarchy is $\lceil \log_c {|\bm{\lambda_o}|} \rceil$. Thanks to the distributed nature of the formation algorithm and the concurrent execution of the agents, the worst-case time complexity of the first stage will be $\mathcal{O}(\log_c {|\bm{\lambda_o}|})$. With the same assumption, it can be easily shown that the resulting hierarchical structure is a complete tree. Hence, denoting the total number of agents in the system by $\mathbb{G}$, this quantity would be:
\begin{equation}
    \frac{c^{\lceil \log_c {|\bm{\lambda_o}|} \rceil}-1}{c-1}<\mathbb{G}\leq \frac{c^{\lceil \log_c {|\bm{\lambda_o}|} \rceil + 1}-1}{c-1}
\end{equation}

\sloppy With that said, the space complexity for the first phase of the proposed technique would be $\mathcal{O}(\frac{c^{\lceil \log_c {|\bm{\lambda_o}|} \rceil + 1}-1}{c-1})=\mathcal{O}(|\bm{\lambda_o}|)$. It is worth noting that among all created agents, only $|\bm{\lambda_o}|$ terminal agents would require dedicated computational resources as they are doing the actual searching and optimization process, and the remaining $\mathbb{G}-|\bm{\lambda_o}|$ can all be hosted and managed together.

The procedures in each round of the second phase of the suggested method can be broken into two main components: (i) transmitting the start coordinates from the root of the hierarchy to the terminal agents, transmitting the results back to the root, and preparing the feedback; and (ii) conducting the actual searching process by the terminal agents to locate a local optimum. The worst case time complexity of preparing the feedback based on the algorithms that were discussed in section~\ref{sec:proposed} would be $\mathcal{O}(|\bm{\lambda_o}|)$, which is because of finding the best candidate among all returned results. In addition, due to the concurrency of the agents, the first component is only processed at the height of the built structure. Therefore, the time complexity of component (i) would be $\mathcal{O}(|\bm{\lambda_o}|+\log_c {|\bm{\lambda_o}|})=\mathcal{O}(|\bm{\lambda_o}|)$. The complexity of the second component, on the other hand, depends on both the budget of the agent, i.e. $b$, and the complexity of building and evaluating the response function $\Psi$. Let $\mathcal{O}(\mathcal{R})$ denote the time complexity of a single evaluation. As a terminal agent makes $b$ number of such evaluations to choose its candidate optima, the time complexity for the agent would be $\mathcal{O}(b\mathcal{R})$. As all agents work in parallel, the complexity of a single iteration at the terminal agents would be $\mathcal{O}(b\mathcal{R})$, leading to the overall time complexity of $\mathcal{O}(|\bm{\lambda_o}|+b\mathcal{R})$. In machine learning problems, we often have $\mathcal{O}(|\bm{\lambda_o}|)\ll\mathcal{O}(\mathcal{R})$. Therefore, if $\mathcal{I}$ denotes the number of iterations until the second phase of the tuning method stops, the complexity of the second stage would be $\mathcal{O}(\mathcal{I} b\mathcal{R})$. The space complexity of the second phase of the tuning method depends on the way that each agent is implementing the main functionalities, such as the learning algorithms they represent, transmitting the coordinates, and providing feedback. Except for the ML algorithms, all internal functionalities of each agent can be implemented using $\mathcal{O}(|\bm{\lambda_o}|)$ space. On the other hand, we have $\mathbb{G}$ agents in the system, which leads to total space complexity of $\mathcal{O}(|\bm{\lambda_o}|^2)$ for non-ML tasks. Let $\mathcal{O}(\mathcal{S})$ denote the worst-case space complexity of a machine learning algorithm that we are tuning. The total space complexity of the second phase of the proposed tuning method would be $\mathcal{O}(|\bm{\lambda_o}|^2+\mathcal{S})$. Similar to the time complexity, in machine learning, we often have $\mathcal{O}(|\bm{\lambda_o}|)\ll\mathcal{O}(\mathcal{S})$, which makes the total space complexity of second phase $\mathcal{O}(\mathcal{S})$. Please note that we have factored out the budgets of the agents and the number of iterations because of not storing the history between different evaluations and iterations.

Considering both stages of the proposed technique, and due to the fact that they are conducted in sequence, the time complexity of the entire steps in an ML hyper-parameter tuning problem, from structure formation to completing the searching operations, would be $\mathcal{O}(\log_c{|\bm{\lambda_o}|}+\mathcal{I} b\mathcal{R})=\mathcal{O}(\mathcal{I} b\mathcal{R})$. Similarly, the space complexity would be $\mathcal{O}(|\bm{\lambda_o}|+\mathcal{S})=\mathcal{O}(\mathcal{S})$.

\subsection{Empirical Results}\label{sec:empi}

This section presents the empirical results of employing the proposed agent-based randomized searching algorithm and discusses the improvements resulting from the suggested inter-agent collaborations. Hyper-parameter tuning in machine learning is basically a back-box optimization problem, and hence, to enrich our empirical discussions, this section also includes results from multiple multi-dimensional optimization problems.

The performance metrics used for the experiments are based on the metrics that are commonly used by the ML and optimization community. Additionally, we analyze the behavior of the suggested methodology based on its own design parameter values, such as budget, width, etc. The methods that have been chosen for the sake of comparison are the standard random search, and the Latin hypercube search methods \cite{bergstra2012random} that are commonly used in practice. Our choices are based on the fact that not only these methods are embarrassingly parallel and are among the top choices to be considered in distributed scenarios, but also they are used as the core optimization mechanisms of the terminal agents in the suggested method and hence can present the impact of the inter-agent collaborations better. In its generic format, as emphasized in \cite{esmaeili2022paams}, one can easily employ alternative searching methods or diversify them at the terminal level as needed.

Throughout the experiments, each terminal agent runs on a separate process, and to make the comparisons fair we keep the number of model/function evaluations fixed among all the experimented methods. To put it in more detail, for a budget value of $b$ for each of $|\bm{\lambda_o}|$ terminal agents and $\mathcal{I}$ number of iterations, the proposed method will evaluate the search space in $b\times\mathcal{I}$ coordinates. We use the same $|\bm{\lambda_o}|$ number of independent agents for the compared random-based methodologies and keeping the evaluation budgets of the agents fixed --- the budgets are assumed to be enforced by the computational limitations of devices or processes running the agents --- we repeat those methods $\mathcal{I}$ times and report the best performance among all agents' repetition history as their final result.

The experiments assess the performance of the proposed method in comparison to the other random-based techniques in 4 categories: (1) iteration-based assessment, which checks the performance of the methods for a particular iteration threshold. In this category, all other parameters, such as budget, connection number, etc. are kept fixed; (2) budget-based assessment, which examines the performance under various evaluation budgets for the terminal agents. It is assumed that all agents have the same budget; (3) width-based assessment, which checks how the proposed method performs for various exploration criteria specified by the slot width parameter; and finally (4) connection-based evaluation, which inspects the effect of the parallel connection numbers that the internal agents can handle. In other words, this evaluation checks if the proposed method is sensitive to the way that the hyper-parameter or decision variables are split during the hierarchy formation phase. All implementations use Python 3.9 and scikit-learn library \cite{scikit-learn}, and the results reported in all experiments are based on 50 different trials.

\begin{figure}
    \includegraphics[width=\textwidth]{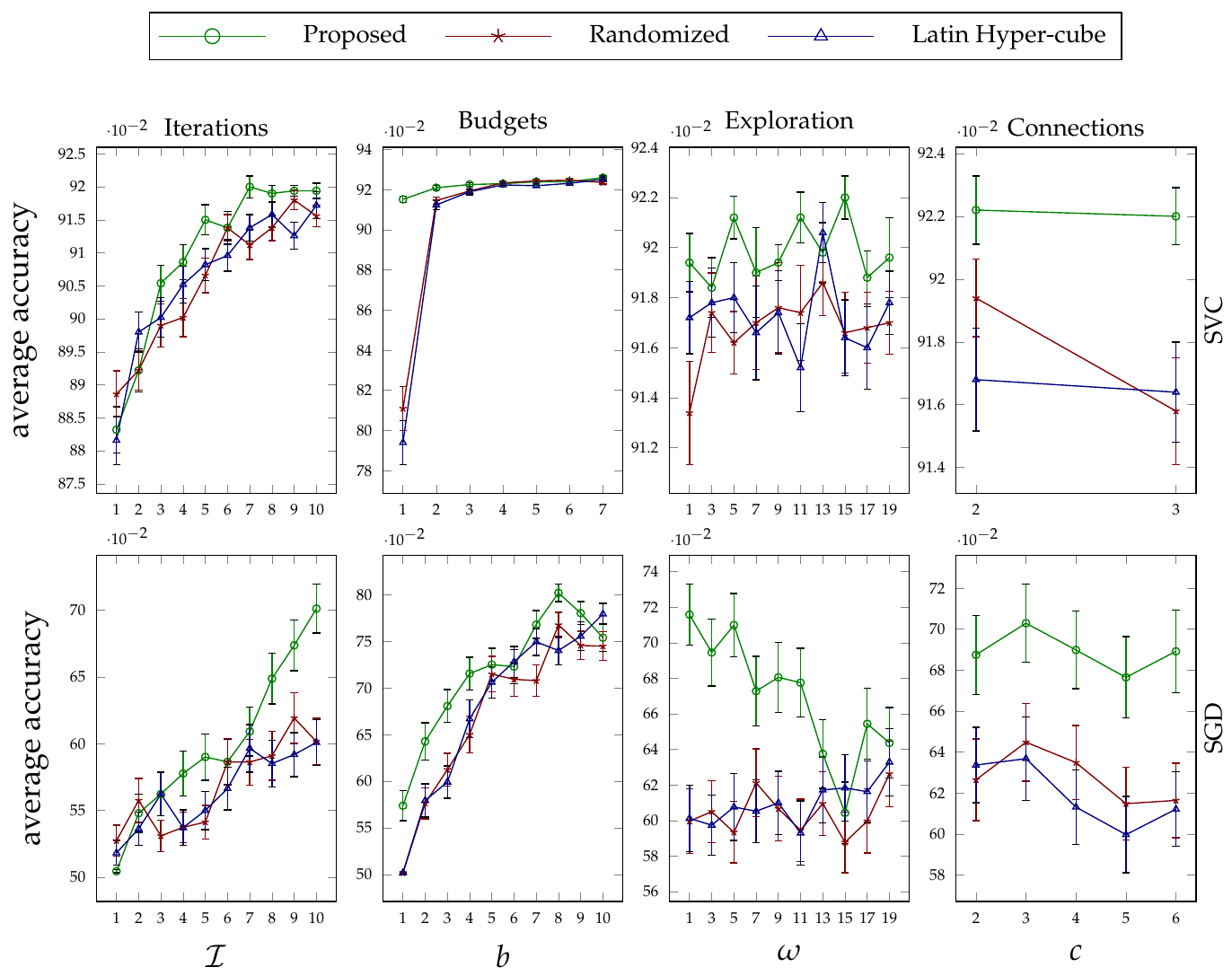}
	\vspace{-3mm}
	\caption{Average performance of C-Support Vector Classification (SVC) (first row) and Stochastic Gradient Descent (SGD) (second row) classifiers on two synthetic classification datasets based on the accuracy measure. The error bars in each plot are calculated based on the standard error.}
	\label{fig:svc-sgd}
\end{figure}

\begin{figure}
    \includegraphics[width=\textwidth]{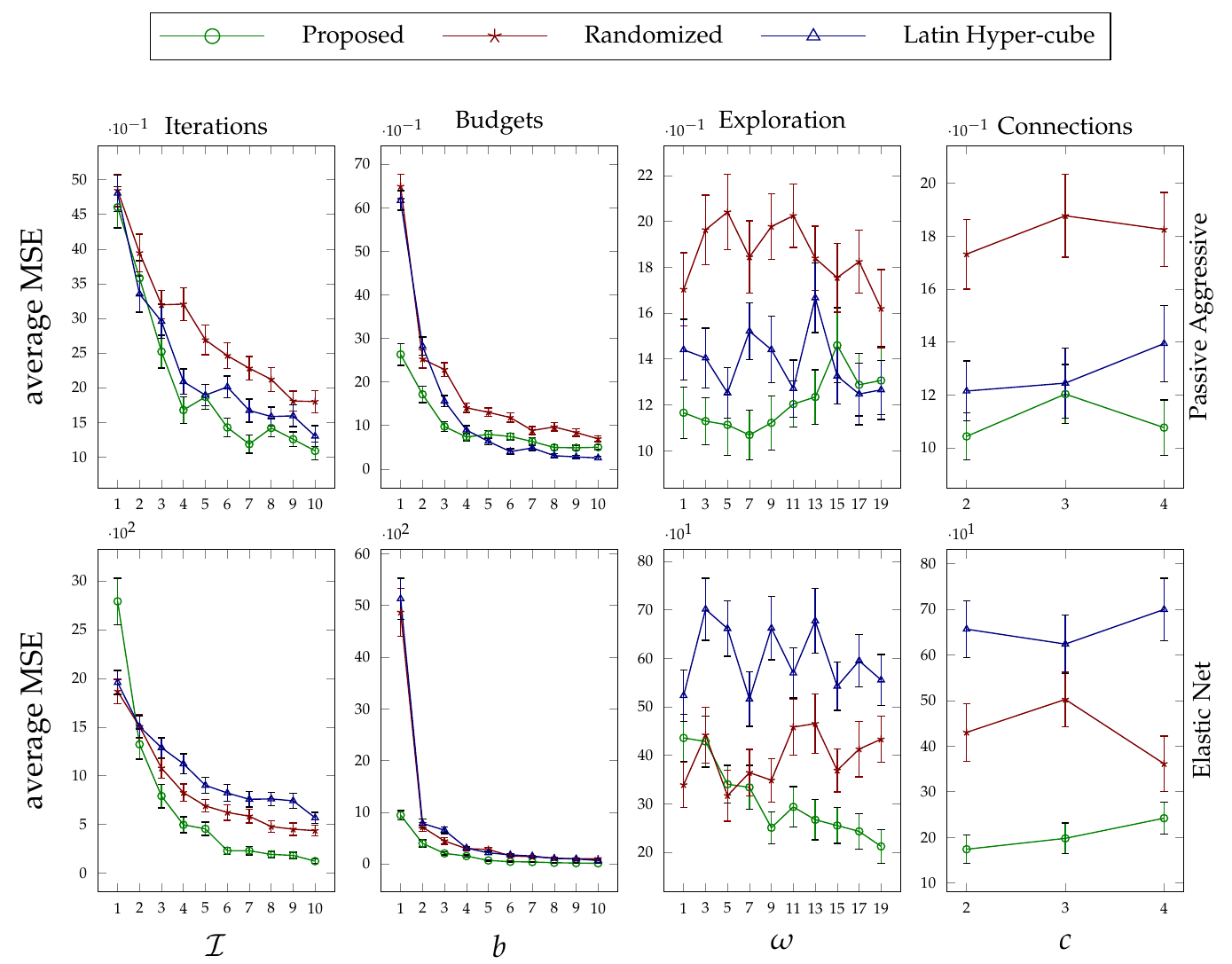}
	\vspace{-3mm}
	\caption{Average performance of Passive Aggressive (first row) and Elastic Net (second row) regression algorithms on two synthetic regression datasets based on the mean squared error (MSE) measure. The error bars in each plot are calculated based on the standard error.}
	\label{fig:pass-elas}
\end{figure}

For the ML hyper-parameter tuning experiments, we have dissected the behavior of the proposed algorithm in two classification and two regression problems. The details of such problems, including the hyper-parameters that are tuned and the used datasets, are presented in table~\ref{tbl:ml-exps}. In all of the ML experiments, we have used 5-fold cross-validation as the model evaluation method. The results obtained for the classification and regression problems are plotted in figures~\ref{fig:svc-sgd} and \ref{fig:pass-elas} respectively.

\begin{table}
    \centering
	\caption{The details of the machine learning algorithms and the datasets used for hyper-parameter tuning experiments. }\label{tbl:ml-exps}
    \resizebox{\textwidth}{!}{
    \begin{threeparttable}
    	\begin{tabular}{l l l l }
    		\toprule
    		ML algorithm &  $\bm{\lambda_o}$ & Dataset & Performance metric \\
    		\midrule
    		C-Support Vector Classification (SVC) \cite{chang2011libsvm,scikit-learn} &  $\{c,\gamma, \text{kernel}\}$\tnote{1} & artificial (100,20)\tnote{$\dagger$} & accuracy \\
    
    		Stochastic Gradient Descent (SGD) Classifier \cite{scikit-learn} &  $\{\alpha,\text{l1\_ratio},\text{tol},\epsilon,\eta_0,\text{val\_frac}\}$\tnote{2} & artificial (500,20)\tnote{$\dagger$} & accuracy \\

            Passive Aggressive Regressor \cite{crammer2006online,scikit-learn} &  $\{c,\text{tol},\epsilon,\text{val\_frac}\}$\tnote{3} & artificial (300,100)\tnote{$\ddagger$} & mean squared error \\

            Elastic Net Regressor \cite{friedman2010regularization,scikit-learn} &  $\{\alpha,\text{l1\_ratio},\text{tol},\text{selection}\}$\tnote{4} & artificial (300,100)\tnote{$\ddagger$} & mean squared error \\
    		\bottomrule
    	\end{tabular}
        \begin{tablenotes}
            \small
            \item[1] $c\sim logUniform(10^{-2},10^{13}), \gamma\sim Uniform(0,1), \text{kernel}\in\{\text{poly, linear, rbf, sigmoid}\}$.
            \item[2] $\alpha\sim Uniform(0,10^{3}), \text{l1\_ratio}\sim Uniform(0,1), \text{tolerance}\sim Uniform(0,10^{3}), \epsilon\sim Uniform(0,10^{3}), \eta_0\sim Uniform(0,10^{3}),\\ \text{validation\_fraction}\sim Uniform(0,1)$.
            \item[3] $c\sim Uniform(0,10^{3}), \text{tolerance}\sim Uniform(0,10^{3}), \text{validation\_fraction}\sim Uniform(0,1), \epsilon\sim Uniform(0,1)$.
            \item[4] $\alpha\sim Uniform(0,1), \text{l1\_ratio}\sim Uniform(0,1), \text{tolerance}\sim Uniform(0,1),\text{selection}\in\{cyclic, random\}$.
            \item[$\dagger$] an artificially generated binary classification dataset using scikit-learn's \texttt{make\_classification} function \cite{scikit-learn-web}. The first number represents the number of samples and the second figure is the number of features.
            \item[$\ddagger$] an artificially generated regression dataset using scikit-learn's \texttt{make\_regression} function \cite{scikit-learn-web}. The first number represents the number of samples and the second figure is the number of features.
        \end{tablenotes}
    \end{threeparttable}
    }
\end{table}

For the \emph{iterations} plot in the first column plots of figures~\ref{fig:svc-sgd} and \ref{fig:pass-elas}, we fixed the parameters of the proposed method for all agents as follows: $b=3, \mathcal{E}=2^{-6}, c=2, \Delta=\{2,2,\dots,2\}$. As can be seen, when the proposed method is allowed to run for more iterations, it yields better performance and its superiority against the other two random-based methods is evident. Comparing the relative performance improvements resulting from the proposed method in the presented ML tasks, it can be seen that as the search space of the agents, and the number of the hyper-parameters needed to be tuned increase, the proposed collaborative method achieved higher improvement. For the SGD classifier, for instance, the objective hyper-parameter set comprises 6 members with continuous domain spaces, and the amount of improvements that have been made after 10 iterations is much higher, about 17\%, than the other experiments with 3-4 hyper-parameters and mixed continuous and discrete domain spaces.

The second column of figures~\ref{fig:svc-sgd} and \ref{fig:pass-elas} illustrates how the performance of the proposed technique changes when we increase the evaluation budgets of the terminal agents. For this set of experiments, we set the parameter values of our method as follows: $\mathcal{I}=10, \mathcal{E}=2^{-6}, c=2, \Delta=\{2,2,\dots,2\}$. By increasing the budget value, the performance of the suggested approach per se improves. However, the rate of improvement slows down for higher budget values, and comparing it against the performance of the other two random-based searching methods, the improvement is significant for lower values of budget. In other words, the proposed tuning method surpasses the other two methods when the agents have limited searching resources. This makes our method a good candidate for tuning the hyper-parameters of deep learning approaches with expensive model evaluations.

The behavior of the suggested method under various exploration parameter values can be seen in column 3 of figures~\ref{fig:svc-sgd} and \ref{fig:pass-elas}. The $\omega$ values on the x-axis of the plots are used to set the initial value for the slot width parameter of all agents using $\mathcal{E}=2^{-\omega-1}$. Based on this configuration, higher values of $\omega$ yield lower values of $\mathcal{E}$, and as a result, more exploitation around the starting coordinates. The other parameters of the method are configured as follows: $\mathcal{I}=10, b=3, c=2, \Delta=\{2,2,\dots,2\}$. Recall from section~\ref{sec:proposed} that the exploration parameter is used by an agent for the dimensions that it does not represent. Based on the results obtained from various tasks, choosing a proper value for this parameter depends on the characteristics of the response function. Having said that, the behavior for a particular task remains almost consistent. Hence, trying one small and one large value for this parameter in a specific problem will reveal its sensitivity and help choose an appropriate value for it.

Finally, the last set of experiments investigates the impact of the number of parallel connections that the internal agents can manage, i.e. $c$, on the performance of the suggested method. The results of this study are plotted on the last column of figures~\ref{fig:svc-sgd} and \ref{fig:pass-elas}. The difference in the number of data points in each plot is because of the difference in the size of the hyper-parameters that we tune in each task. The values of the parameters that we kept fixed for this set of experiments are as follows: $\mathcal{I}=10, b=3, \mathcal{E}=2^{-6}, \Delta=\{2,2,\dots,2\}$. As can be seen from the illustrated results, the proposed method is not very sensitive to the value that we choose or is enforced by the system for parameter $c$. This parameter plays a critical role in the shape of the hierarchy that is distributedly formed in phase 1 of the suggested approach; therefore, one can opt to choose a value that fits with the connection or computational resources that are available without sacrificing the performance very much.

As stated before, we have also studied the suggested technique for the global optimization problem to see how it performs in finding the optima of various convex and non-convex functions. These experiments also help us closely check the relative performance improvements in higher dimensions. We have chosen three non-convex benchmark optimization functions and a convex toy function the details of which are presented in table~\ref{tbl:func-exps}. For each function, we run the experiments in three different dimension sizes, and the goal of the optimization is to find the global minimum. Very similar to the settings that we discussed for ML hyper-parameter tuning, whenever we mean to fix the value of each parameter values in different experiment sets, we use the following parameter values: $\mathcal{I}=10, b=3, \mathcal{E}=2^{-10}, c=2, \Delta=\{2,2,\dots,2\}$.

\begin{table}
    \centering
	\caption{The details of the multi-dimensional functions used for global optimization experiments. }\label{tbl:func-exps}
    \resizebox{\textwidth}{!}{
    \begin{threeparttable}
    	\begin{tabular}{l l l l }
    		\toprule
    		Function &  $\bm{\lambda_o}$ & Domain &$f(\bm{x}^*)$ \\
    		\midrule
    		Hartmann, 3D, 4D, 6D \cite{Jamil_2013} &  $\{x_1,\dots, x_d\}, d\in\{3,4,6\}$&$x_i\in\left[0,1\right]$ & \textbf{3D:}-3.86278, \textbf{4D:}-3.135474, \textbf{6D:}-3.32237\\

            Rastrigin, 3D, 6D, 10D \cite{rudolph1990globale} &  $\{x_1,\dots, x_d\}, d\in\{3,6,10\}$&$x_i\in\left[-5.12,5.12\right]$  & \textbf{3D:}0,\textbf{6D:}0, \textbf{10D:}0\\

            Styblinski-Tang, 3D, 6D, 10D \cite{styblinski1990experiments} &  $\{x_1,\dots, x_d\}, d\in\{3,6,10\}$ &$x_i\in\left[-5,5\right]$ & \textbf{3D:}-117.4979,\textbf{6D:}-234.9959, \textbf{10D:}391.6599\\

            Mean Average Error, 3D, 6D, 10D\tnote{$\dagger$} &  $\{x_1,\dots, x_d\}, d\in\{3,6,10\}$&$x_i\in\left[0,100\right]$  & \textbf{3D:}0,\textbf{6D:}0, \textbf{10D:}0\tnote{$\ddagger$}\\    
    		
    		\bottomrule
    	\end{tabular}
        \begin{tablenotes}
            \item[$\dagger$] This is a toy multi-dimensional MAE function that is defined as $f(\bm{x})=\frac{1}{n}\sum_{i=1}^{n}{|\bm{x}-\chi|}$ where $\chi$ denotes a ground truth vector that is generated randomly in the domain space for each experiment. 
            \item[$\ddagger$] This is a convex function and the coordinate of its minimum value depends on the ground truth vector that is generated, i.e. when $\bm{x}=\chi$.
        \end{tablenotes}
    \end{threeparttable}
    }
\end{table}

\begin{figure}
    \includegraphics[width=\textwidth]{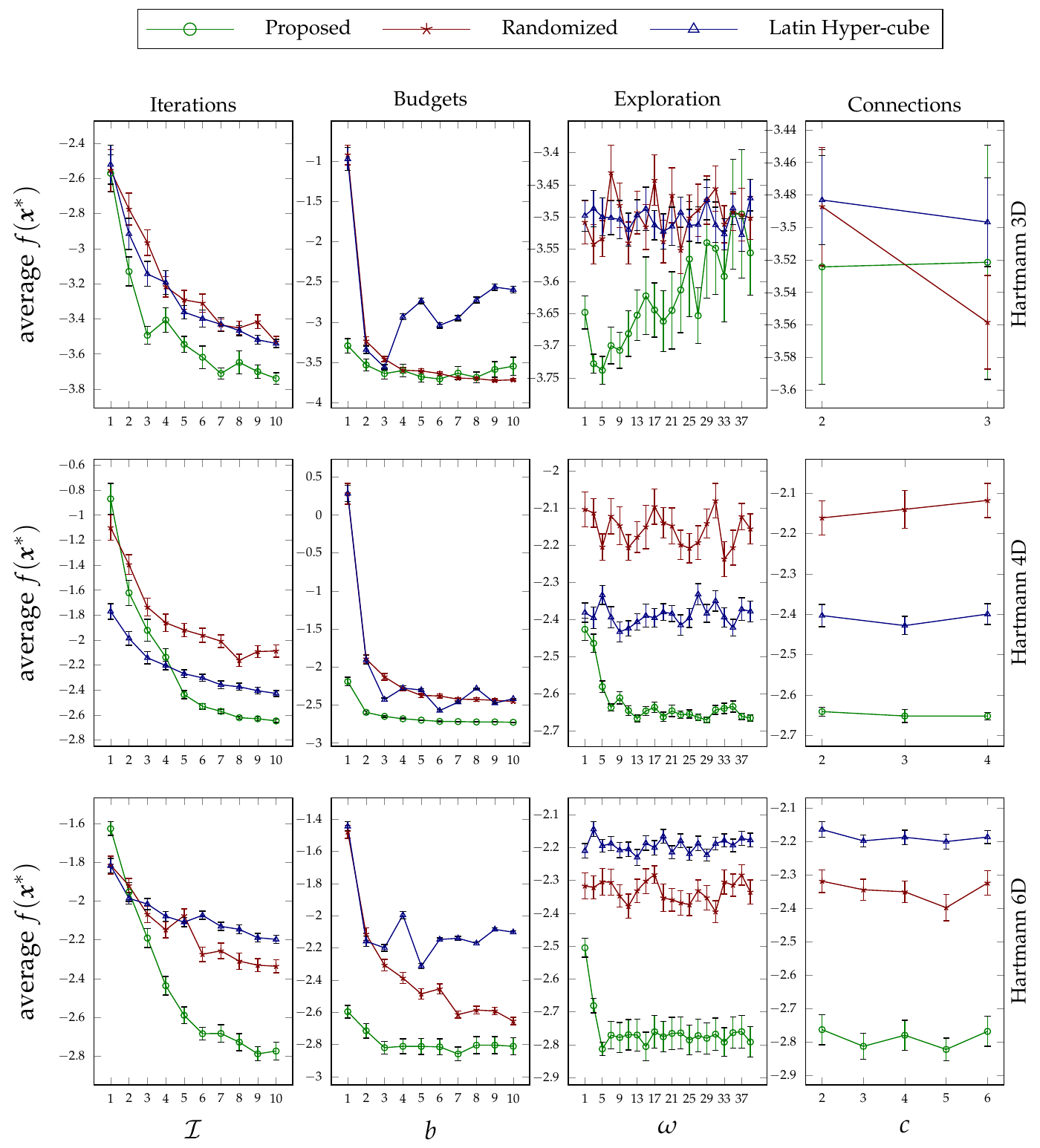}
	\vspace{-3mm}
	\caption{Average function values of the Hartmann function under variable iteration, budget, exploration, and connection thresholds. Each row of the figure pertains to a particular dimension size and the error bars are calculated based on the standard error.}
	\label{fig:hartmann}
\end{figure}

\begin{figure}
    \includegraphics[width=\textwidth]{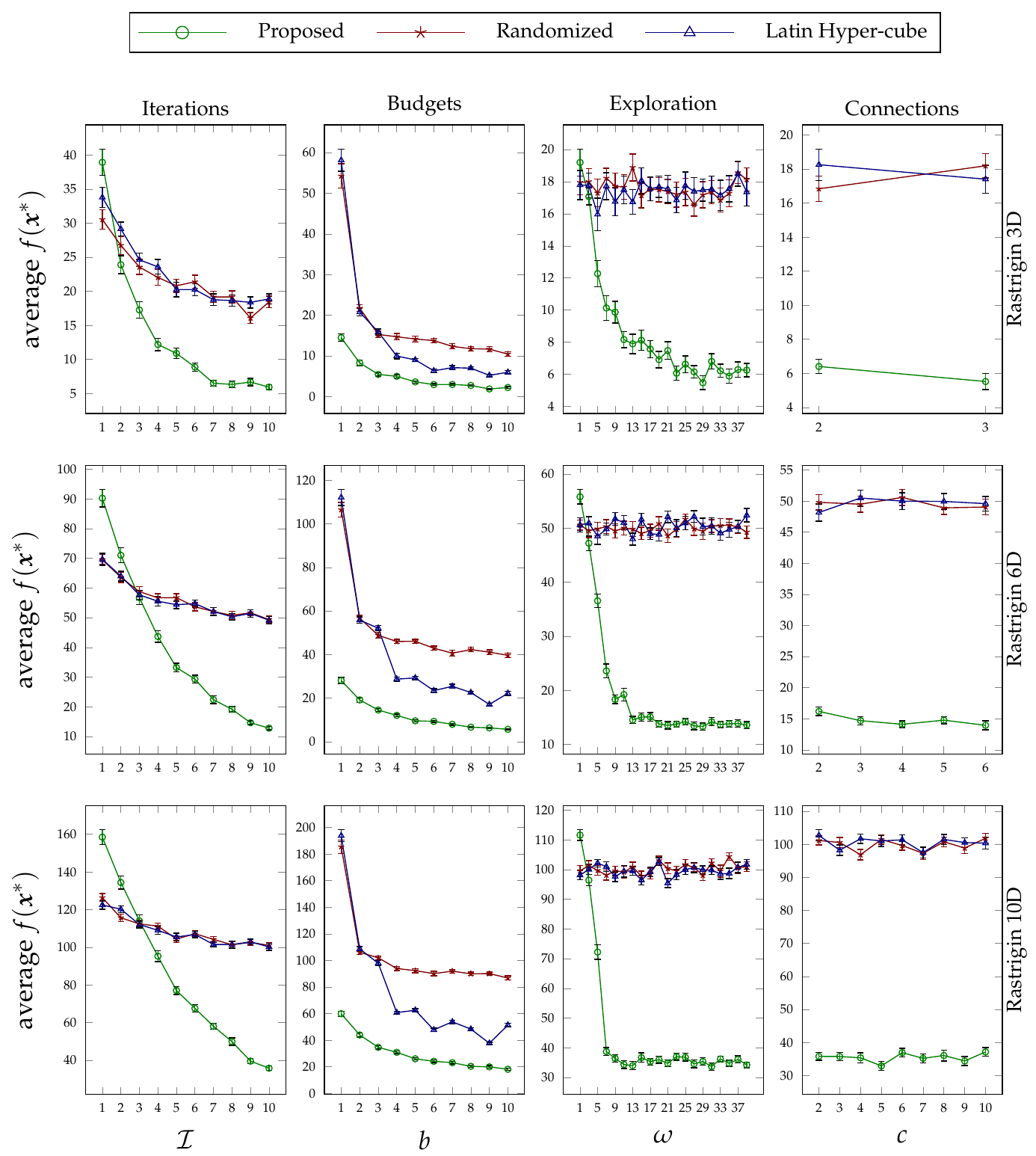}
	\vspace{-3mm}
	\caption{Average function values of the Rastrigin function under variable iteration, budget, exploration, and connection thresholds. Each row of the figure pertains to a particular dimension size and the error bars are calculated based on the standard error.}
	\label{fig:rastrigin}
\end{figure}

\begin{figure}
    \includegraphics[width=\textwidth]{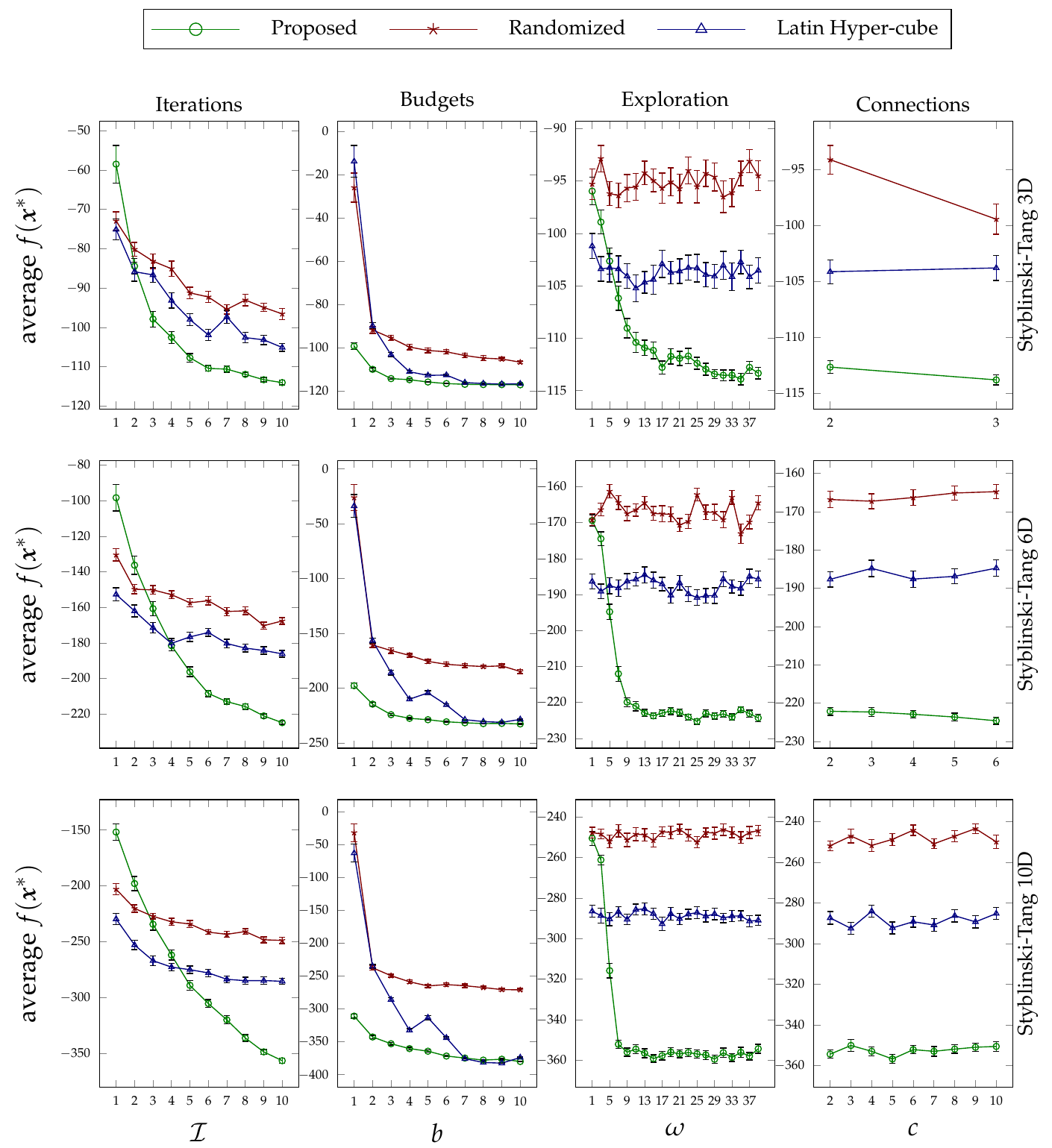}
	\vspace{-3mm}
	\caption{Average function values of the Styblinski-Tang function under variable iteration, budget, exploration, and connection thresholds. Each row of the figure pertains to a particular dimension size and the error bars are calculated based on the standard error.}
	\label{fig:styblinskitang}
\end{figure}

\begin{figure}
    \includegraphics[width=\textwidth]{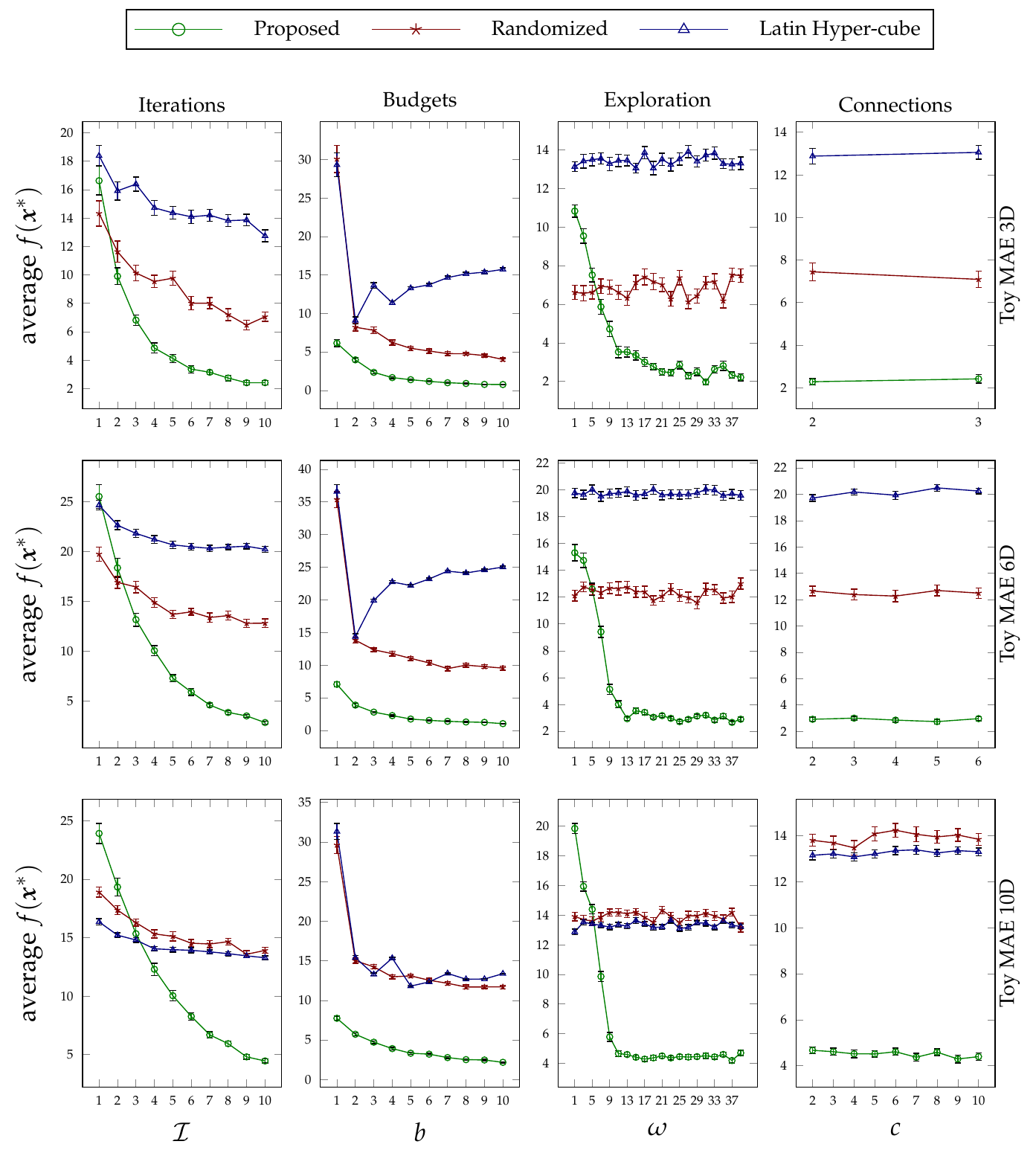}
	\vspace{-3mm}
	\caption{Average function values of the Toy Mean Absolute Error function under variable iteration, budget, exploration, and connection thresholds. Each row of the figure pertains to a particular dimension size and the error bars are calculated based on the standard error.}
	\label{fig:toymae}
\end{figure}

The plots are grouped by functions and can be found in figures~\ref{fig:hartmann}, \ref{fig:rastrigin}, \ref{fig:styblinskitang}, and \ref{fig:toymae}. The conclusion that was made about the behavior of the proposed approach under different values of its design parameters applies in these optimization experiments as well. That is, the more the proposed method runs, the better performance it achieves; its superiority on low budget values is clear; its sensitivity to exploration parameter values is consistent; and the way that the decision variables set is broken down during the formation of the hierarchy does not affect the performance very much. Furthermore, as can be seen in each group figure, the proposed algorithm yields a better minimum point in comparison to the other two random-based methods when the dimensionality of a function increases.

It is worth emphasizing that the contribution of this paper is not to compete with the state-of-the-art in function optimization but to propose a distributed tuning/optimization approach that can be deployed on a set of distributed and networked devices. The discussed analytical and empirical results not only demonstrated the behavior and impact of the design parameters that we have used in our approach but also suggested the way that they can be adjusted for different needs. We believe the contributed approach of this paper can be significantly improved provided with more sophisticated and carefully chosen tuning strategies and corresponding configurations.   
    \section{Conclusion}\label{sec:conclusion}
This paper presented an agent-based collaborative random search method that can be used for machine learning hyper-parameter tuning and global optimization problems. The approach employs two types of agents during the tuning/optimization process: the internals and the terminal agents that are responsible for facilitating the collaborations and tuning an individual decision variable respectively. Such agents and the interaction network between them are created during the hierarchy formation phase and remain the same for the entire run-time of the suggested method. Thanks to the modular and distributed nature of the approach and its procedures, it can be easily deployed on a network of devices with various computational capabilities. Furthermore, the design parameters used in this technique enable each individual agent to customize its own searching process and behavior independent from its peers in the hierarchy, and hence, allow employing diversity in both algorithmic and deployment levels.

The paper dissected the proposed model from different aspects and provided some tips on handling its behavior for various applications. According to the analytical discussions, our approach requires slightly more amount of computational and storage resources than the traditional and Latin hyper-cube randomized search methods that are commonly used for both hyper-parameter tuning and black-box optimization problems. However, this results in significant performance improvements, especially in computationally restricted circumstances and higher numbers of decision variables. This conclusion was verified in both machine learning model tuning tasks and general multi-dimensional function optimization problems. 

The presented work can be further extended both technically and empirically. As was discussed throughout this paper, we kept the searching strategies and the way the design parameters are configured as simple as possible so we could reach a better understanding of the effectiveness of collaborations and searching space divisions. A few potential extensions in this direction include: utilizing diverse searching methods and hence having a heterogeneous multi-agent system at the terminal level; splitting the searching space not based on the dimensions but based on the range of the values that decision variables in each dimension can have; employing more sophisticated collaboration techniques; and using a learning-based approach to dynamically adapt the values of the design parameters during the runtime of the method. Empirically, on the other hand, the presented research can be extended by applying it to expensive machine learning tasks such as tuning deep learning models with a large number of hyper-parameters. We are currently working on some of these studies and suggest them as future work. 

    \bibliographystyle{unsrt}
	\bibliography{main}
	
\end{document}